\documentclass[sigconf,natbib=false,nonacm]{acmart}
\AtBeginDocument{%
  }

\setcopyright{acmlicensed}

\copyrightyear{2026}
\acmYear{2026}
\acmConference[SAC '26]{The 41st ACM/SIGAPP Symposium on Applied Computing}{March 23--27, 2026}{Thessaloniki, Greece}
\acmBooktitle{The 41st ACM/SIGAPP Symposium on Applied Computing (SAC '26), March 23--27, 2026, Thessaloniki, Greece}
\acmPrice{}

\RequirePackage[
  datamodel=acmdatamodel,
  style=acmnumeric,
  ]{biblatex}

\graphicspath{{figures/}}
\usepackage{pifont}
\usepackage{tikz}
\usepackage{url} 

\begin{document}

\title{Learning to Detect Baked Goods with Limited Supervision}
\renewcommand{\shorttitle}{Detecting Baked Goods with Limited Labels}


\author{Thomas H. Schmitt}
\affiliation{%
  \institution{Technische Hochschule Nürnberg Georg Simon Ohm}
  \city{Nuremberg}
  \country{Germany}
}
\email{thomas.schmitt@th-nuernberg.de}

\author{Maximilian Bundscherer}
\affiliation{%
  \institution{Technische Hochschule Nürnberg Georg Simon Ohm}
  \city{Nuremberg}
  \country{Germany}
}

\author{Tobias Bocklet}
\affiliation{%
  \institution{Technische Hochschule Nürnberg Georg Simon Ohm}
  \city{Nuremberg}
  \country{Germany}
}




\renewcommand{\shortauthors}{T. Schmitt et al.}
\begin{abstract}
Monitoring leftover products provides valuable insights that can be used to optimize future production.
This is especially important for German bakeries because freshly baked goods have a very short shelf life.
Automating this process can reduce labor costs, improve accuracy, and streamline operations.
We propose automating this process using an object detection model to identify baked goods from images.
However, the large diversity of German baked goods makes fully supervised training prohibitively expensive and limits scalability.
Although open-vocabulary detectors (e.g., OWLv2, Grounding DINO) offer flexibility, we demonstrate that they are insufficient for our task.
While motivated by bakeries, our work addresses the broader challenges of deploying computer vision in industries, where tasks are specialized and annotated datasets are scarce.
We compile dataset splits with  varying supervision levels, covering \(19\) classes of baked goods.
We propose two training workflows to train an object detection model with limited supervision.
First, we combine OWLv2 and Grounding DINO localization with image-level supervision to train the model in a weakly supervised manner.
Second, we improve viewpoint robustness by fine-tuning on video frames annotated using Segment Anything 2 as a pseudo-label propagation model.
Using these workflows, we train YOLOv11 for our detection task due to its favorable speed-accuracy tradeoff.
Relying solely on image-level supervision, the model achieves a mean Average Precision (mAP) of \(0.91\).
Fine-tuning with pseudo-labels raises model performance by \(19.3\%\) under non-ideal deployment conditions.
Combining these workflows trains a model that surpasses our fully-supervised baseline model under non-ideal deployment conditions, despite relying only on image-level supervision.
\end{abstract}

\ccsdesc[500]{Applied computing~Industry and manufacturing}
\ccsdesc[300]{Computing methodologies~Machine learning}
\ccsdesc[300]{Computing methodologies~Computer vision}
\ccsdesc[300]{Computing methodologies~Semi-supervised learning settings}
\keywords{
food recognition,
baked goods,
computer vision,
object detection,
weakly supervised learning,
pseudo-labeling
}

\maketitle
\renewcommand\thefootnote{} 
\footnote{%
© ACM 2026. This is the author's version of the work. It is posted here for personal use. Not for redistribution. 
The definitive Version of Record was published in The 41st ACM/SIGAPP Symposium on Applied Computing (SAC '26), 
\url{http://dx.doi.org/10.1145/3748522.3779800}}%
\addtocounter{footnote}{-1} 
\section{Introduction}
\label{sec:intro}
In the food industry, optimizing production is vital for both economic and environmental reasons, as it bolsters resource efficiency and reduces food waste.
This is especially consequential for German bakeries, where the limited shelf life of freshly baked goods amplifies the economic and environmental impact of overproduction.
To reduce waste, German bakeries commonly reprocess leftovers into breadcrumbs or repurpose them as animal feed.
However, these practices do not address the underlying mismatch between production planning and actual demand.
Monitoring leftover baked goods provides essential feedback for production optimization.
Because manual counting is labor-intensive, bakeries 
often forgo this step to reduce labor costs.
Automating counting with a machine learning model can provide bakeries with more timely feedback while reducing manual labor costs.
%

%
During reprocessing, leftover baked goods are arranged flat on metal drying sheets.
This arrangement creates favorable conditions for computer vision object detection models to automatically detect and count them.
However, due to the wide variety of baked goods in Germany, there are few public datasets with the necessary, fine-grained annotations.
Prior works that adapt computer vision for the food industry \cite{bread01, bread02} primarily focus on more generic tasks and train models in a fully supervised manner.
Moreover, because each bakery’s assortment is unique, adapting models typically requires bakery-specific fine-tuning.
Although dataset collection is feasible, manual annotation remains the critical bottleneck for model adoption, as it is both costly and time-consuming.
This application exemplifies the broader challenges of adapting computer vision models for industrial deployment: models must be tailored to highly specific tasks; yet, compiling annotated datasets is prohibitively expensive.
%

%
In this work, we demonstrate how to train an object detection model with a fine-grained understanding of baked goods with limited supervision.
Prior works on computer vision models in data-scarce scenarios have focused on architectures that enable few-shot learning or allow zero-shot inference \cite{zero-shot, GLIP, OWL, DE-Vit, G-DINO, OWLv2, Yolo-world}.
However, such models are typically evaluated on broad, more generic benchmark datasets.
We test whether open-vocabulary object detection models, OWLv2 \cite{OWLv2} and Grounding DINO \cite{G-DINO}, can be effectively applied to our task without additional training.
We combine these models with image-level supervision to generate bounding box annotations in a weakly supervised manner.
To improve model robustness, we employ Segment Anything 2 \cite{SAM2} as a pseudo-label propagation model, generating annotations with minimal or no manual supervision.
We train the one-stage object detection model YOLOv11 \cite{YOLOv11} to detect baked goods, as its favorable speed-accuracy tradeoff makes it well-suited for on-device deployment.
To train our model, we compile dataset splits with varying levels of supervision (partially and weakly annotated) covering \(19\) classes of baked goods.
To evaluate our models, we compiled a fully annotated application-relevant dataset comprising \(763\) images.
We compare models trained with limited supervision against a strong fully supervised baseline and show that they can nearly match — and in some cases surpass — its performance.
The main contributions of this work are:
\begin{itemize}
\item A scalable, cost-efficient workflow for training object detection models with fine-grained domain understanding, integrating open-vocabulary detectors with image-level supervision.
\item A training workflow that improves model robustness via efficient pseudo-label propagation, requiring minimal or no manual supervision.
\item Evaluation of our training workflows on a real-world, app\-lication-relevant dataset against a strong baseline, demonstrating both competitive performance and practical relevance.
\end{itemize}
\section{Related Work}
\label{sec:related_work}
Prior works adapting computer vision models for the food industry typically target less specific tasks under relatively controlled deployment conditions.
Most works rely on fully supervised training, which requires fully annotated datasets and thereby limits scalability and practical adoption.
In contrast, our work aims to reduce the amount of supervision needed during training and fine-tuning.
%

%
Schmitt et al. \cite{semmel01, semmel02} used computer vision models to detect and count baked goods in images, aiming to support production optimization.
They enhanced training data with the Copy-Paste augmentation \cite{CopyPaste} and image-to-image translation models CycleGAN \cite{CycleGAN} and pix2pix \cite{pix2pix} to increase appearance diversity and simulate occlusions.
However, their models were evaluated under idealized conditions and were not integrated into a bakery's workflow.
Their best model achieved an mAP of \(0.90\) at an IoU threshold of \(0.5\).
Kılcı et al. \cite{bread01} applied computer vision models to automate biscuit quality inspection.
They trained YOLOv8 \cite{YOLOv8} in a fully supervised manner to identify and classify biscuit defects, thereby supporting hygienic and consistent production.
They compiled a dataset of \(4,\!990\) biscuit images captured under fixed lighting and camera conditions.
Their best binary classification model (defect vs. non-defect) achieved an accuracy of \(96.78\%\).
Raja et al. \cite{bread02} applied computer vision models to automate bread quality inspection.
They combined thermal imaging with object detection models from the YOLO-series \cite{YOLOv5, YOLOv8, YOLOv9, YOLOv11} to non-destructively detect contaminants in bread.
They compiled a dataset of \(650\) thermal images of bread with various contaminations.
Although the dataset contained multiple contamination types, the model was trained only to localize, not classify, contaminations.
Their best model achieved a mAP of \(0.607\).
Their pruned model optimized for real-time inference achieved a mAP of \(0.601\).
%

%
Prior works that train models with limited supervision, or design architectures adaptable with little or no supervision, typically evaluate their models on standard benchmark datasets \cite{COCO, PascalVoc, LVIS}.
Even works that use more relevant benchmarks often evaluate on comparably generic datasets.
In contrast, we adapt our models to a distinct, fine-grained food dataset using limited supervision.
Moreover, we compare our models against a strong baseline trained in a fully supervised manner.
%

%
Open-vocabulary object detection models \cite{GLIP, OWL, DE-Vit, G-DINO, OWLv2, Yolo-world} can be applied without task-specific fine-tuning.
However, these models rely heavily on pretrained text encoders \cite{CLIP, BERT} to encode task semantics.
In our detection task, the link between baked good names and their appearance is questionable, particularly when culinary authenticity and specificity are lost in translation.
Hongxu Ma et al. \cite{zero-shot} proposed an object detection model for fine-grained zero-shot detection, without relying on a pretrained text encoder.
To evaluate their model, they collected the FGZSD-Birds dataset, which comprises \(148,\!820\) images spanning \(1,\!432\) bird species.
Their model achieved a mAP of \(0.785\) at an IoU threshold of \(0.5\).
However, despite being designed for zero-shot detection, their model was pretrained on the bird domain using a training split of FGZSD-Birds.
\section{Data}
\label{sec:data}
\begin{figure}
	\centering
    \includegraphics[width=\linewidth, trim=0pt 0pt 0pt 0pt, clip]{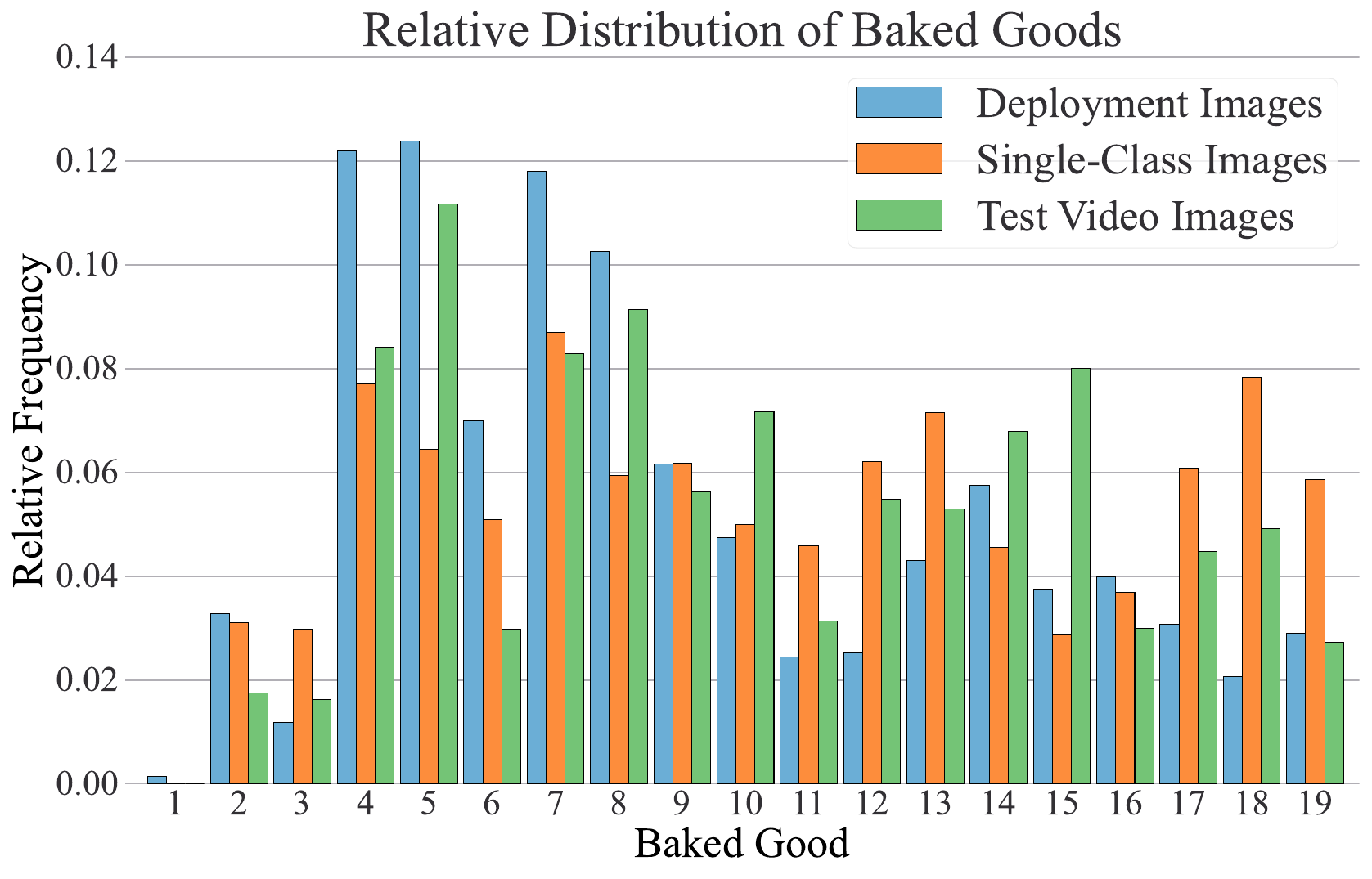}
    \caption{Relative class distributions of baked goods across Deployment (\(D\)), Single-Class (\(C_{\text{train}}\)) and Test Video (\(V_{\text{test}}\)) Images.}
	\label{fig:histo_data}
\end{figure}

We collected our datasets in collaboration with a local bakery, which provided access to its products and allocated staff time for data collection.
Our datasets were primarily gathered using a layman-friendly iOS application under realistic deployment conditions (see Section~\ref{sec:data_dep_env}).
Images are captured during leftover reprocessing, either immediately before they are turned into breadcrumbs or repurposed as animal feed.
Our datasets distinguish between \(19\) classes of baked goods (\(18\) named products and one fall-back class).
These classes form a representative subset of our collaborating bakery’s product range. 
While the bakery produces more than \(18\) distinct baked goods, the remaining products are of deferred practical relevance, and their availability varies seasonally.
Table~\ref{tab:names} lists the German names of the baked goods along with their English translations.
The original German names are retained to preserve culinary authenticity and specificity.
\begin{table}
    \centering
    \small
    \setlength{\tabcolsep}{6pt} 
    \begin{tabular}{l|l}
        \toprule
        German Name                       & English Translation \\
        \midrule
        \midrule
        Backware                          & Backware \\
        Bauernbrot                        & Farmer's Bread \\
        Flößerbrot                        & Raftsman's Bread \\
        Salzstange                        & Pretzel Stick \\
        Sonnenblumensemmel                & Sunflower Seed Roll \\
        Kürbiskernsemmel                  & Pumpkin Seed Roll \\
        Roggensemmel                      & Rye Bread Roll \\
        Dinkelsemmel                      & Spelt Bread Roll \\
        Laugenstange Schinken-Käse        & Ham and Cheese Pretzel Stick \\
        Pfefferlaugenbrezel               & Pepper Pretzel \\
        Kernige Schinken-Käse-Stange      & Hearty Ham and Cheese Stick \\
        Schokocroissant                   & Chocolate Croissant \\
        Apfeltasche                       & Apple Turnover \\
        Quarktasche                       & Quark Cheese Pastry \\
        Mohnschnecke                      & Poppy Seed Swirl Pastry \\
        Nussschnecke                      & Nut Swirl Pastry \\
        Vanillehörnchen                   & Vanilla Croissant \\
        Kirschtasche                      & Cherry Turnover \\        
        Früchteschiffchen Erdbeere        & Strawberry Fruit Tart \\
        \bottomrule
    \end{tabular}
    \caption{German names of the \(19\) baked good classes, retained for culinary authenticity, alongside their English translations.}
    \label{tab:names}
\end{table}
Figure~\ref{fig:histo_data} shows the relative distributions of baked goods in our datasets.
The first class, \textit{Backware} (German for \textit{baked good}), serves as a fallback class for miscellaneous or unforeseen baked goods.
Despite its high heterogeneity, this fallback class contributes only marginally to our datasets, accounting for fewer than \(0.1\%\) of all object instances.
Table~\ref{tab:data} summarizes our dataset splits, their notation, and key statistics.
\begin{table}
    \begin{center}
        \small
        \setlength{\tabcolsep}{6pt} 
        \begin{tabular}{l|r|r|r|l}
            \toprule
            Dataset & \#Images & \#Instances & \#Videos & Purpose \\
            \midrule
            \midrule        
            \(D_{\text{train}}\) & 610       & 12261         & \multicolumn{1}{c|}{-} & Baseline train \\
            \(D_{\text{test}}\)  & 153       & 3075          & \multicolumn{1}{c|}{-} & Benchmark test \\
            \(C_{\text{train}}\) & 315       & 2220        & \multicolumn{1}{c|}{-} & Weakly-sup train \\
            \(V_{\text{train}}\) & 4,945     & 499445        & 167                    & Pseudo-ft \\
            \(V_{\text{test}}\)  & 1,186     & 119786        & 42                     & Viewpoint test \\
            \bottomrule
        \end{tabular}
        \caption{Overview of our dataset splits, showing sizes, modalities, and purposes.}
        \label{tab:data}
    \end{center}
\end{table}
\subsection{Deployment Conditions}
\label{sec:data_dep_env}
We integrate our trained models and image-capturing setup into an iOS application, enabling direct process integration and efficient on-device inference.
The iOS application is designed for layman operators, allowing them to use our models and capture relevant images with minimal training.
To facilitate model performance and ensure reliable integration, we define our deployment conditions with the following constraints:
(1) Images are captured by a fixed HD webcam mounted directly overhead to ensure image consistency and optimal visibility of the baked goods.
(2) Before images are captured, baked goods are laid flat on a metal drying tray, a routine step in the breadcrumb reprocessing process.
(3) While ambient lighting in the bakery remains fairly consistent, we added an external light source to further stabilize the lighting conditions.
This overhead setup introduces bias that limits model generalizability.
However, it can readily be enforced in active production using a simple image-capturing rig.
\subsection{Deployment Images}
\label{sec:data_dep_imgs}
\begin{figure}
	\centering
    \includegraphics[width=\linewidth, trim=100pt 0pt 100pt 0pt, clip]{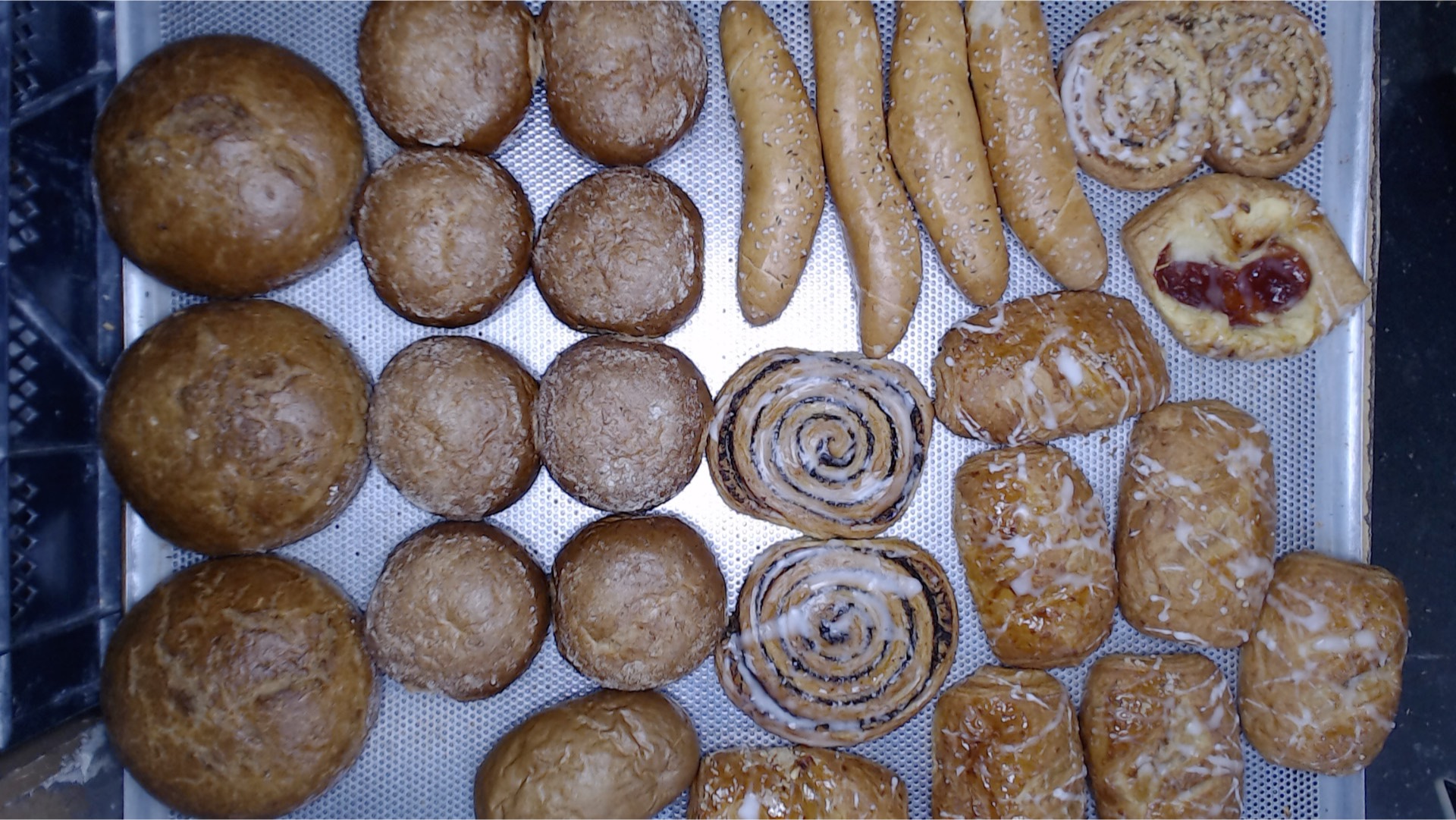}
    \caption{Sample deployment image captured by bakery staff using the iOS application.}
	\label{fig:D_imgs}
\end{figure}

To train our baseline model and evaluate model performance, we compiled an application-relevant deployment image dataset (\(D\)).
The images were captured directly by a bakery employee using our iOS application to ensure dataset realism.
To preserve image authenticity, the employee was instructed to arrange baked goods on the drying tray following the bakery’s standard procedure.
Images were collected over a six-month period to capture both intra-product variance and seasonal variations in the baked goods.
In total, we collected \(763\) images, with an average of approximately \(20\) baked goods per image.
Figure~\ref{fig:D_imgs} shows a sample deployment image.
%

%
Using the LabelStudio tool~\cite{LabelStudio}, we manually annotated the images with bounding boxes.
Annotating each image required approximately three minutes, yielding a total annotation time of roughly 
\(38\) hours.
Given this annotation effort, training models in a fully supervised manner for active production would be both time-consuming and prohibitively expensive.
This motivates our adoption of weak and pseudo supervision for model training.
We adopt an \(80/20\) \(\text{train}/\text{test}\) split, yielding a baseline training set (\(D_{\text{train}}\)) with \(610\) images and a deployment test set (\(D_{\text{test}}\)) with \(153\) images.
\subsection{Single-Class Images}
\label{sec:data_class_imgs}
To train our models in a semi-supervised manner, we compiled a single-class image dataset (\(C_{\text{train}}\)).
To form this dataset, employees were instructed not to mix baked goods on the drying trays. 
This restriction allows us to infer the class of the baked goods  in each image using image-level supervision at minimal annotation cost.
Object localization is automated using zero-shot object detection models (see Section~\ref{sec:methods_weak_ann}).
We collected single-class images for each of our \(18\) non-fallback baked good classes.
In total, we collected \(315\) single-class images, with an average of \(7.0\) baked goods per image.
%

%
Each baked good class was represented by at least \(12\) instances.
Due to seasonality, \textit{Kernige Schinken-Käse-Stange} and \textit{Nussschnecke} had the fewest representatives in our single-class dataset, with \(12\) each.
To offset availability imbalances, baked goods with fewer representatives were over-sampled during image capturing.
\subsection{Video Frames}
\label{sec:data_vid_imgs}
To assess and improve model robustness to viewpoint variations, we compiled a video dataset (\(V\)) using a setup that mimics our deployment conditions.
While our models are trained on individual frames, recording videos and converting them into image sequences streamlines the collection process and enables efficient pseudo-label propagation.
In total, we collected \(209\) videos of baked goods, with an average duration of about \(7\) seconds each.
Each video features an average of about \(10\) baked goods, prominently visible in the initial frames.
Each baked good class is represented by at least \(7\) representatives.
Each video begins with a top-down view, consistent with the perspective of our deployment images \(D\) and single-class images \(C\).
During recording, the camera is progressively tilted from a top-down view to slanted angles to increase detection difficulty.
%

%
We adopt an \(80/20\) \(\text{train}/\text{test}\) split, yielding \(167\) training and \(42\) test videos.
At a rate of \(4\) fps, we extracted \(4,\!945\) training set (\(V_{\text{train}}\)) and \(1,\!186\) test set (\(V_{\text{test}}\)) frames from our videos.
Each resulting test set image was manually annotated with bounding boxes.
To reduce annotation costs, only the first frames of each video in the training set were manually annotated with bounding boxes.
The remaining annotations were generated automatically using pseudo-label propagation (see Section~\ref{sec:data_dep_env}).
This pseudo-labeling reduced the manual annotation cost by more than \(96\%\).
\section{Annotation Workflows}
\label{sec:methods}
\subsection{Weakly Supervised Image Annotation}
\label{sec:methods_weak_ann}
OWLv2 \cite{OWLv2} and Grounding DINO \cite{G-DINO} are open-vocabulary object detection models that, in principle, are capable of detecting any object based on user-provided text queries.
However, their inference cost is computationally prohibitive, especially for on-device deployment.
Moreover, their effectiveness on our specialized cross-domain task, which requires fine-grained domain understanding, is uncertain.
We, therefore, primarily leverage them as zero-shot localization models.
Given the single-class structure of dataset \(C_{\text{train}}\), we supplement these localizations with image-level supervision to automatically generate bounding box annotations in a weakly supervised manner.
Since a baked good's name does not reliably describe its appearance, we avoid class-specific prompting.
Instead, we prompt OWLv2 and Grounding DINO with the generic text query \textit{baked good} to localize baked goods in \(C_{\text{train}}\).
%

\begin{figure}
\centering
\includegraphics[width=\linewidth, trim=200pt 0pt 0pt 0pt, clip]{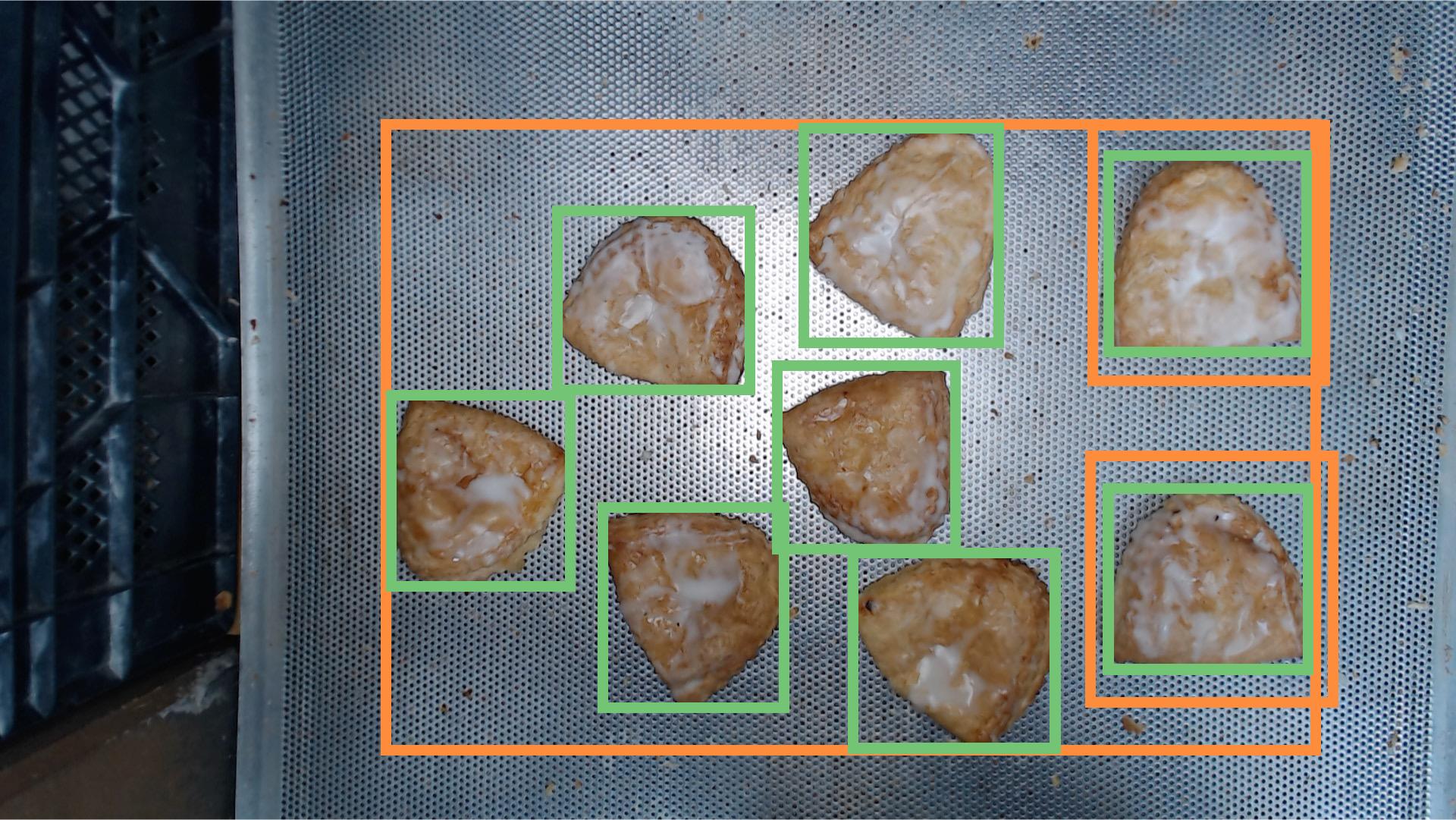}
\caption{Effect of our post-processing on a Grounding DINO prediction for a single-class image containing the baked good \textit{Apfel\-tasche}. Filtered bounding boxes are shown in orange.}
\label{fig:posprocessing}
\end{figure}

%
In preliminary experiments, we observed that OWLv2 and Grou\-nding DINO consistently over-predict objects in our \(C_{\text{train}}\) images.
To mitigate this over-prediction, we introduce four sequential post-processing steps:
(1) \textit{Background Filter}: removes background predictions by discarding bounding boxes covering at least \(90\%\) of the image area.
(2) \textit{Duplicate Filter}: removes redundant predictions by discarding lower-confidence bounding boxes that overlap with other bounding boxes at an IoU exceeding \(0.75\).
(3) \textit{Crowd Filter}: removes bounding boxes likely containing multiple objects by discarding any bounding box that contains three or more other bounding boxes.
We define a bounding box as contained within another if its capture rate (intersection over bounding box area) exceeds \(0.95\).
(4) \textit{Nested Filter}: removes partial-object predictions by discarding bounding boxes contained within another bounding box (capture rate exceeding \(0.95\)).
Filter thresholds were determined heuristically to optimize annotation quality.
Figure~\ref{fig:posprocessing} shows the effect of our post-processing on a Grounding DINO prediction.
\subsection{Pseudo-Labeling via Object Tracking}
\label{sec:methods_pseudo_ann}
Segment Anything 2 and its predecessor, Segment Anything \cite{SAM}, are promptable segmentation models with state-of-the-art zero-shot generalization capabilities. 
However, adapting them to a highly specialized detection task presents several hurdles.
(1) Their model architectures result in prohibitively high inference costs, especially for on-device deployment.
(2) When operating in a zero-shot manner, they predict only the segmentation masks and no object classes.
Consequently, (3) without input queries, they cannot distinguish background regions from task-relevant objects.
%

%
We bypass these hurdles by using Segment Anything 2 as a pseudo-label propagation model to accelerate the annotation of our sequential image datasets \(V_{\text{train}}\) and \(V_{\text{test}}\).
To track the baked goods, the model is initialized with bounding box queries for the first frame images of each video.
These first frame bounding box queries can be created manually or generated automatically by a pretrained detection model.
\section{Experiments and Results}
\label{sec:experiments}
We train the one-stage object detection model YOLOv11 \cite{YOLOv11} on our detection task.
To train the model, we use three supervision regimes: fully supervised, weakly supervised (Section~\ref{sec:exp_train_on_weak}), and pseudo-supervised fine-tuning (Section~\ref{sec:train_on_pseudo}). 
We chose YOLOv11 due to its competitive performance and comparably low computational cost, making it well-suited for on-device inference.
We adopted the largest variant, YOLOv11x (\(58.1\text{M}\) parameters) trained at an input resolution of \(1280\times1280\text{px}\) to maximize detection fidelity.
On our deployment device (iPad Mini 6), this model achieves a mean latency of under \(0.5\) seconds per image.
This latency is acceptable for workflow integration, as bakery employees tolerate sub-second delays without requiring real-time responsiveness.
Unless stated otherwise, we fine-tune YOLOv11x pretrained on the COCO dataset \cite{COCO} to facilitate model performance.
%

%
Models are trained for \(100\) epochs using stochastic gradient descent (SGD) with momentum and cosine learning rate decay to ensure convergence.
To suppress duplicate object detection and better preserve object counts, we adopt class-agnostic non-maximum suppression (NMS).
The remaining model and training hyperparameters are set to their respective default values.
We primarily report model performance using COCO-style mean Average Precision (mAP@[0.5:0.95]), computed in \(0.05\) IoU increments, subsequently referred to as mAP.
To assess the statistical significance of performance differences, we use a proportion z-test \cite{z-test} at a significance level of \(p=0.05\).
\subsection{Weakly Supervised Image Annotation}
\label{sec:exp_weak_ann}
We evaluate OWLv2 and Grounding DINO as zero-shot localization models for baked goods.
Each model is prompted with the generic text query \textit{baked good} to localize each baked good in an image.
Given the image structure of our single-class dataset \(C_{\text{train}}\), we can supplement these localizations with class labels via image-level supervision to fully annotate it in a weakly supervised manner.
We evaluate the effectiveness of our proposed post-processing steps (Section~\ref{sec:methods_weak_ann}).
YOLOv11x, trained on our baseline training set \(D_{\text{train}}\), serves as our baseline annotation model.
Table~\ref{tab:zero-shot_ann} shows the resulting annotation performance.

\begin{table}
    \begin{center}
        \small
        \setlength{\tabcolsep}{6pt} 
        \begin{tabular}{l|r|r|r|r|r}
        \toprule
        Ann Model & Query & Post & mAP & \(P\) & \(R\) \\
        \midrule
        \midrule
        OWLv2          & translation            & \ding{51}              & 0.03 & \multicolumn{1}{c|}{-} & \multicolumn{1}{c}{-} \\
        G-DINO         & translation            & \ding{51}              & 0.02 & \multicolumn{1}{c|}{-} & \multicolumn{1}{c}{-} \\
        YOLOv11x       & \multicolumn{1}{c|}{-} & \multicolumn{1}{c|}{-} & 1.00 & 1.00 & 1.00 \\
        OWLv2          & baked good             & \ding{55}              & 0.94 & 0.84 & 1.00 \\
        OWLv2          & baked good             & \ding{51}              & 0.99 & 0.99 & 1.00 \\
        G-DINO         & baked good             & \ding{55}              & 0.71 & 0.66 & 1.00 \\
        G-DINO         & baked good             & \ding{51}              & 0.99 & 1.00 & 1.00 \\
        \bottomrule
        \end{tabular}
        \caption{Zero-shot performance of OWLv2 and Grounding DINO (G-DINO) on \(C_{\text{train}}\), using either text query \textit{baked good} for localization or the translation for detection, with/without post-processing (Post). Presented are the mean Average Precision (mAP), along with Precision (\(P\)) and Recall (\(R\)) calculated at an IoU threshold of \(0.75\). YOLOv11x, trained on \(D_{\text{train}}\) (YOLOv11x) serves as our baseline.}
        \label{tab:zero-shot_ann}
    \end{center}
\end{table}
To justify using OWLv2 and Grounding DINO as localization models, we evaluate their zero-shot baked goods detection performance on \(C_{\text{train}}\).
For each baked good, the models are prompted to detect it using its English name (see Table~\ref{tab:names}).
The resulting annotation performances are shown in Table~\ref{tab:zero-shot_ann}.
To assess the quality of our translations, we compute the top-5 average cosine similarity between their CLIP \cite{CLIP} embeddings.
A high value (close to \(1.00\)) would indicate that the embedding model struggles to differentiate the translations, making it more difficult for the models to separate the corresponding object concepts.
The embeddings of our translations have a top-5 average cosine similarity of \(0.78\).
To contextualize this value, the embeddings of the \(80\) COCO class names have a top-5 average cosine similarity of \(0.836\).
This suggests that CLIP, the text embedding model used by both OWLv2 and Grounding DINO, effectively captures the semantic differences in the embedding space.
However, OWLv2 and Grounding DINO fail to differentiate between classes of baked goods based on our translations.
This is likely due to the domain gap between their training data and our detection task, compounded by the weak correlation between a baked good’s appearance and its name.
Ignoring misclassifications between baked goods, OWLv2 and Grounding DINO achieve a class-agnostic Average Precision (aAP) of \(0.97\), indicating that the main challenge lies in distinguishing between baked goods rather than detecting them, which motivated our approach.
%

%
OWLv2 and Grounding DINO are effective detection models for our baked goods when prompted with text query \textit{baked good}.
Our post-processing steps significantly improve annotation quality.
Grounding DINO predictions especially benefit from our post-processing steps, resulting in annotations that surpass those of OWLv2 and are comparable to those predicted by our baseline annotation model.
After post-processing, Grounding DINO annotations contain only \(11\) false positives and \(5\) false negatives, which we consider negligible and attributable to noise.
These results demonstrate that, when combined with our post-processing steps, Grounding DINO exhibits minimal mislocalization or hallucinations on our \(C_{\text{train}}\) images.
\subsection{Training on Weakly Annotated Images}
\label{sec:exp_train_on_weak}
We evaluate whether model performance can be retained when we train on our smaller training set \(C_{\text{train}}\), annotated in a weakly supervised manner.
To this end, we train YOLOv11x on \(C_{\text{train}}\), using either ground truth or predicted annotations.
YOLOv11x, trained on our baseline training set \(D_{\text{train}}\), serves as our baseline model.
Table~\ref{tab:train_zero-shot} shows the resulting model performances.
\begin{table}
    \begin{center}
        \small
        \setlength{\tabcolsep}{6pt} 
        \begin{tabular}{l|r|r|r}
        \toprule
        Train Set                & Ann Model                      & mAP & cAP [min, max]\\
        \midrule
        \midrule
        \(D_{\text{train}}\)     & \multicolumn{1}{c|}{-}         & 0.98 & \(\left[0.988,\,0.995\right]\) \\
        \(C_{\text{train}}\)     & GT                             & 0.90 & \(\left[0.811,\,0.995\right]\) \\
        \(C_{\text{train}}\)     & OWLv2                          & 0.89 & \(\left[0.691,\,0.995\right]\) \\
        \(C_{\text{train}}\)     & G-DINO                         & 0.91 & \(\left[0.853,\,0.993\right]\) \\      
        \bottomrule
        \end{tabular}        
        \caption{Model performance on \(D_{\text{test}}\) with \(C_{\text{train}}\) annotations, either taken from the ground truth (GT) or predicted using OWLv2 or Grounding DINO (G-DINO). Presented are the mAP and the range of class-wise Average Precisions (cAP), excluding our fallback class.}
        \label{tab:train_zero-shot}
    \end{center}
\end{table}
%

%
We retain a mAP of \(0.90\) compared to our baseline of \(0.98\) on \(D_{\text{train}}\), while training solely on \(C_{\text{train}}\).
Our models suffer no significant performance drop when trained on \(C_{\text{train}}\) annotated in a weakly supervised manner.
Therefore, and based on the results presented in Section~\ref{sec:exp_weak_ann}, subsequent experiments are conducted using \(C_{\text{train}}\) annotated by Grounding DINO.
The performance drop relative to our baseline model is primarily due to the increased variance in class-wise Average Precisions \(cAP\).
While our baseline model detects each class of baked good with roughly equal accuracy, even our best model, trained on \(C_{\text{train}}\), struggles to accurately detect certain baked goods.
It specifically struggles to detect baked goods \textit{Kernige Schinken-Käse-Stange} and \textit{Nussschnecke}, for which it achieves a \(cAP\) of \(0.85\) and \(0.91\), respectively.
The poor detection coincides with the number of object instances representing these baked goods in \(C_{\text{train}}\), as they were the two least-represented (see Section~\ref{sec:data_class_imgs}).
Ignoring these poorly detected baked goods, the model achieves a mAP of \(0.97\) on our deployment test set images \(D_{\text{test}}\).
The remaining performance gap is due to the reduced intra-product variability of \(C_{\text{train}}\) relative to \(D_{\text{train}}\), since it features less than half as many baked-good instances.
\subsection{Pseudo-Labeling via Object Tracking}
\label{sec:pseudo_ann}
We evaluate the effectiveness of Segment Anything 2 as a pseudo-labeling model for our video frame datasets derived from our video dataset (\(V\)).
To this end, we track baked goods in \(V_{\text{test}}\) using first-frame bounding box queries, sourced either from the ground truth or predicted by a trained YOLOv11x model.
When bounding box queries are predicted, we discard bounding boxes with confidence scores below \(0.7\) to suppress false positives.
Alongside the tracking mAP on our test set \(V_{\text{test}}\), we report the performance of our query models on the first frames of each video (\(\text{mAP}_{F1}\)).
Table~\ref{tab:object_tracking} shows the resulting annotation performances.
\begin{table}
    \begin{center}
        \small
        \setlength{\tabcolsep}{6pt} 
        \begin{tabular}{l|r|r|r}
        \toprule
        F1-Ann     & \(\text{mAP}_{F1}\) &  \(\text{mAP}_{T}\) & \(\text{aAP}_{T}\) \\
        \midrule
        \midrule
        GT &         \multicolumn{1}{c|}{-}      & 1.00 & 1.00 \\
        Trained on \(D_{\text{train}}\)  & 0.95  & 0.95 & 0.98 \\
        Trained on \(C_{\text{train}}\)  & 0.80  & 0.78 & 0.95 \\
        \bottomrule
        \end{tabular}
        \caption{Annotation Performance on \(V_{\text{test}}\) using the Segment Anything 2 object tracking model, where first-frame bounding box queries (F1-Ann) are either taken from the ground truth (GT) or predicted by YOLOv11x, trained on \(D_{\text{train}}\) or \(C_{\text{train}}\). Presented are the mAP on the first frames (\(\text{mAP}_{F1}\)), on \(V_{\text{test}}\) (\(\text{mAP}_{T}\)), and the class-agnostic Average Precision aAP on \(V_{\text{test}}\) (\(\text{aAP}_{T}\)).}
        \label{tab:object_tracking}
    \end{center}
\end{table}
%

%
Segment Anything 2 perfectly tracks all baked goods in \(V_{\text{test}}\) when the first-frame bounding box queries are taken from the ground truth.
When first-frame prompts are predicted by a trained YOLOv11x model, tracking performance depends on the quality of those predicted bounding boxes.
These results demonstrate that Segment Anything 2 achieves robust tracking on our \(V_{\text{test}}\) video sequences, with no observable drift or compounding errors.
Using YOLOv11x trained either on \(D_{\text{train}}\) or \(C_{\text{train}}\) to predict the first-frame bounding-box queries, we achieve a mAP of \(0.95\) and \(0.78\) on \(V_{\text{test}}\), respectively.
However, most errors stem from object misclassifications, with tracking based on model predictions achieving an aAP of \(0.98\) and \(0.95\) on \(V_{\text{test}}\), respectively.
\subsection{Training on Pseudo-Labeled Video Frames}
\label{sec:train_on_pseudo}
We evaluate whether we can adapt YOLOv11x to detect baked goods from arbitrary camera angles using training data annotated via pseudo-labeling.
To this end, we fine-tune our pretrained YOLOv11x models on pseudo-annotated \(V_{\text{train}}\), supplemented with \(C_{\text{train}}\) or \(D_{\text{train}}\).
Annotations for \(V_{\text{train}}\) are predicted via Segment Anything 2 object tracking, using first-frame bounding box queries either taken from the ground truth or predicted by a trained YOLOv11x model.
Alongside the mAP on the full test set \(V_{\text{test}}\), we report the performance of our models on the initial frames of each video (\(\text{mAP}_{FI}\)).
YOLOv11x, trained either on \(C_{\text{train}}\) or \(D_{\text{train}}\), serve as our baseline models.
Table~\ref{tab:train_object_tracking} shows the resulting model performances.
\begin{table}
    \begin{center}
        \small
        \setlength{\tabcolsep}{6pt} 
        \begin{tabular}{l|r|r|r}
        \toprule
        Train Sets     & F1-Ann     & \(\text{mAP}_{FI}\) & \(\text{mAP}_{T}\) \\
        \midrule
        \midrule
        \(D_{\text{train}}\)                           & \multicolumn{1}{c|}{-}            & 0.97  & 0.83 \\
        \(C_{\text{train}}\)                           & \multicolumn{1}{c|}{-}            & 0.88  & 0.63 \\
        \(D_{\text{train}}\) + \(V_{\text{train}}\)    & GT                                & 0.99  & 0.99 \\
        \(D_{\text{train}}\) + \(V_{\text{train}}\)    & Trained on \(D_{\text{train}}\)   & 0.99  & 0.99 \\      
        \(C_{\text{train}}\) + \(V_{\text{train}}\)    & Trained on \(C_{\text{train}}\)   & 0.89  & 0.88 \\  
        \bottomrule
        \end{tabular}
        \caption{Model performance on \(V_{\text{test}}\) when training on different training sets (Train set), and using different first-frame annotation models (F1-Ann). Presented are the mAP on the initial frames of each video (\(\text{mAP}_{FI}\)) and on \(V_{\text{test}}\).}
        \label{tab:train_object_tracking}
    \end{center}
\end{table}
%

%
Our baseline models perform significantly worse on \(V_{\text{test}}\) than on \(D_{\text{test}}\) (see Table~\ref{tab:train_zero-shot}).
This suggests a lack of model robustness to viewpoint variations.
Their detection performance on the initial frames closely mirrors that on \(D_{\text{test}}\), indicating that the initial frames in \(V_{\text{test}}\) are visually similar to the images in \(D_{\text{test}}\).
%

%
Fine-tuning models on \(V_{\text{train}}\) significantly boosts model performance on \(V_{\text{test}}\).
Moreover, model performance remains consistent throughout \(V_{\text{test}}\), showing no significant decline on the entire videos compared to just the initial frames.
Notably, models suffer no significant performance drop when the first-frame annotations are predicted by YOLOv11x, trained on \(D_{\text{train}}\).
When first-frame annotations are predicted with YOLOv11x, trained on \(C_{\text{train}}\), model performance drops significantly, consistent with its weaker performance on \(D_{\text{test}}\) and the initial frames of \(V_{\text{test}}\).
Notably, however, YOLOv11x pretrained on \(C_{\text{train}}\) and fine-tuned on \(V_{\text{train}} + C_{\text{train}}\), with first-frame annotations predicted by YOLOv11x trained on \(C_{\text{train}}\), outperforms both baselines on \(V_{\text{test}}\) despite relying solely on weak supervision.
\subsection{Effect of Camera Angle}
\label{sec:exp_angle}
\begin{figure}
	\centering
    \includegraphics[width=\linewidth]{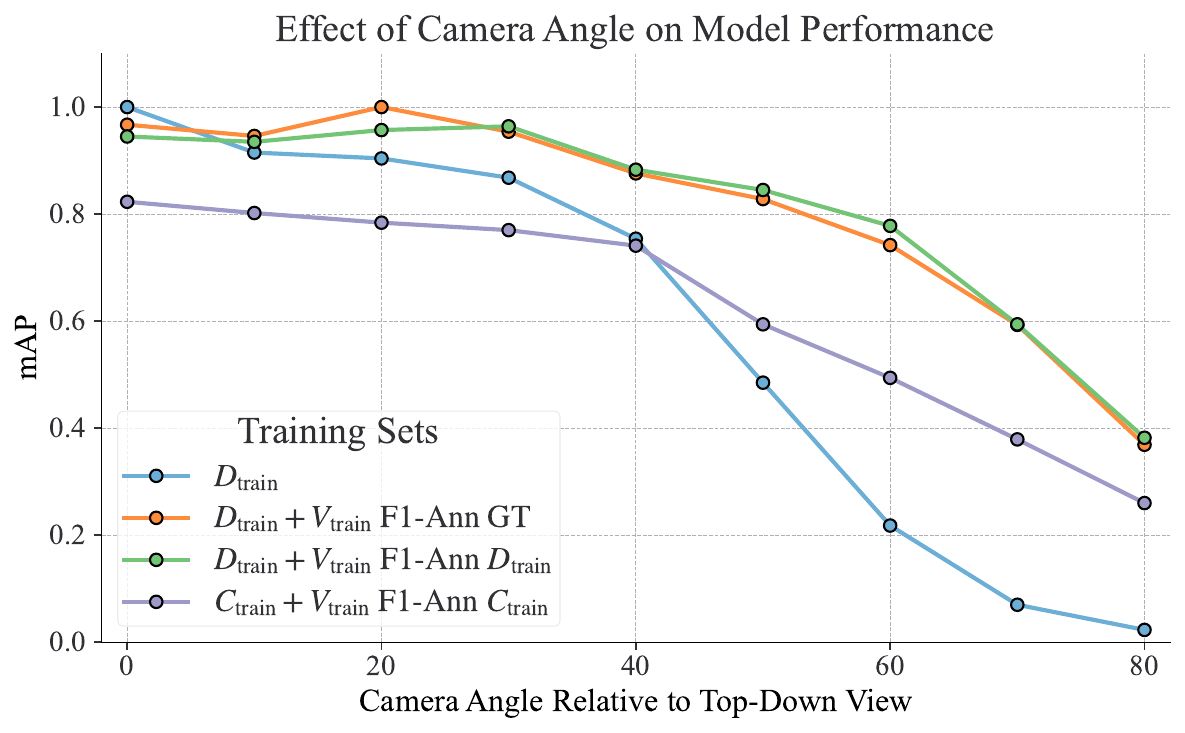}
    \caption{Model mAP on \(A_{\text{test}}\) as a function of the relative camera angle, illustrating the effect of viewpoint variation on detection performance.}
	\label{fig:effect_camera_angle}
\end{figure}

Fine-tuning models on \(V_{\text{train}}\) significantly boosts their resilience to viewpoint variations.
However, the results in Section~\ref{sec:train_on_pseudo} do not directly reveal how the camera angle affects model performance.
To directly analyze this aspect, we define a test set \(A_{\text{test}}\), specifically designed to isolate this variable.
\(A_{\text{test}}\) comprises \(90\) images of baked goods captured under conditions resembling our deployment conditions.
Images were captured at fixed camera angles ranging from \(0^\circ\) to \(80^\circ\) in \(10^\circ\) increments relative to the top-down view.
In total, we captured \(10\) images per angle, each manually annotated with bounding boxes.
Despite significant inter-object occlusions at camera angles beyond \(60^\circ\), we ensured that each baked good remained distinctly visible.
We use this dataset to evaluate the robustness of our trained models across varying camera angles.
YOLOv11x, trained on \(D_{\text{train}}\), serves as our baseline model.
Figure~\ref{fig:effect_camera_angle} shows the model mAP as a function of camera angle.
%

\begin{figure}
	\centering
    \includegraphics[width=\linewidth, trim=350pt 250pt 370pt 250pt, clip]{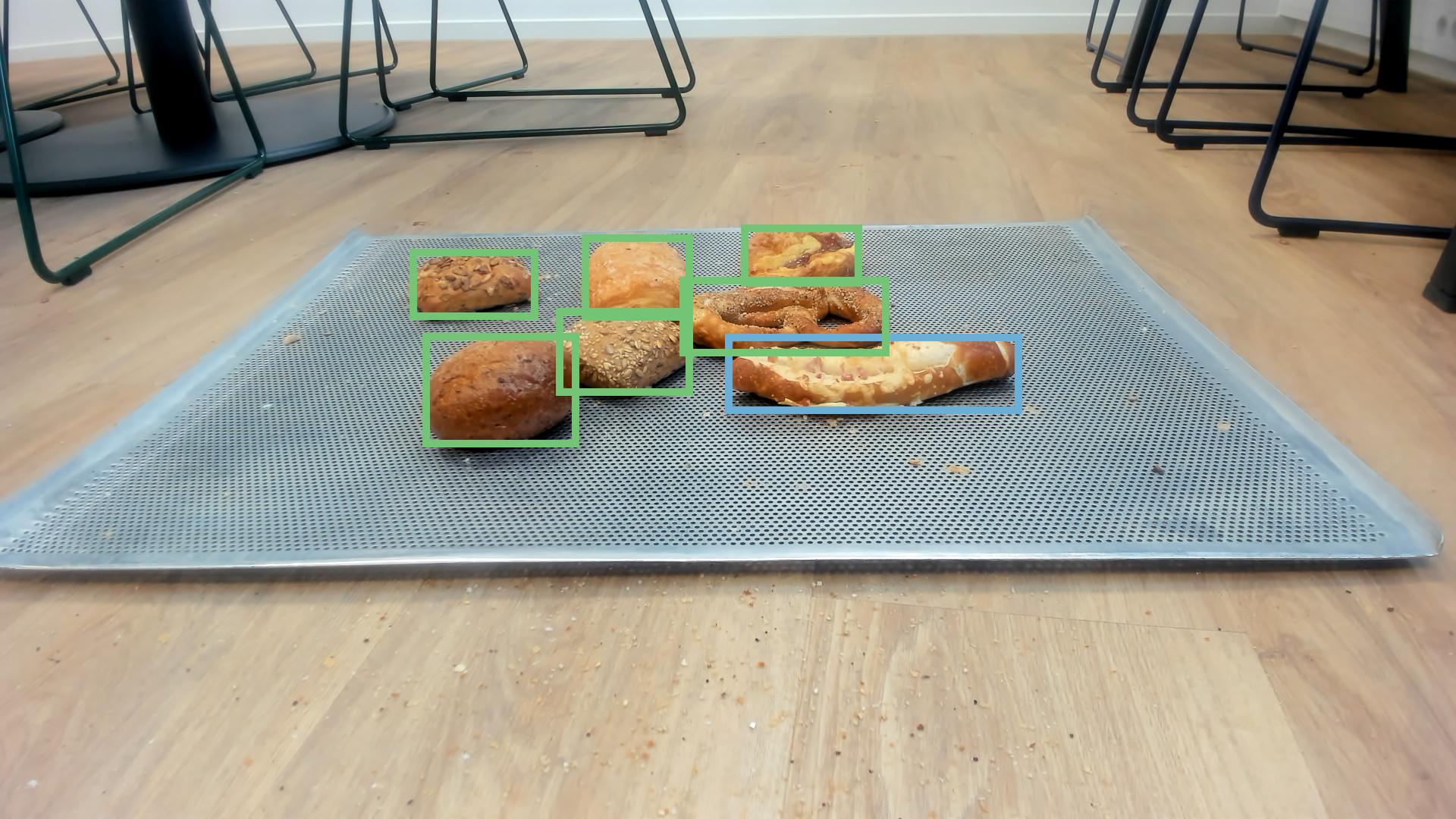}
    \caption{Predictions on an \(A_{\text{test}}\) image captured at a \(60^\circ\) relative to the top-down view. Green bounding boxes mark correct detections by the fine-tuned model; blue boxes mark detections correctly made by both the fine-tuned and baseline model.}
	\label{fig:img_60deg}
\end{figure}

%
All models, trained on \(D_{\text{train}}\), achieve a relatively stable mAP exceeding \(0.90\) up to a camera angle of \(20^\circ\).
Model performance deteriorates once the camera angle exceeds \(30^\circ\), with especially sharp declines beyond \(40^\circ\).
The performance of our baseline model drops sharply, reaching an mAP as low as \(0.22\) at a \(60^\circ\) camera angle.
While both models, trained on \(V_{\text{train}}\) and \(D_{\text{train}}\), also experience significant performance drops beyond \(30^\circ\), they retain a mAP exceeding \(0.70\) up to a camera angle of \(60^\circ\).
If the camera angle exceeds \(60^\circ\), their mAP likewise drops to as low as \(0.37\).
However, at such steep camera angles, object detection becomes ambiguous and unreliable.
This ambiguity arises from both optical distortions and severe or complete inter-object occlusion occurring at steep camera angles.
There is no significant performance difference on \(A_{\text{test}}\) whether first-frame annotations are taken from the ground truth or predicted.
YOLOv11x, trained on \(C_{\text{train}}\) and \(V_{\text{train}}\), performs worse than our baseline on \(A_{\text{test}}\) images taken at shallow relative camera angles.
However, it exhibits greater robustness to changing camera angles.
Notably, it surpasses our baseline model at camera angles exceeding \(40^\circ\), despite relying solely on weak supervision.
Figure~\ref{fig:img_60deg} shows the predictions of YOLOv11x, fine-tuned on \(V_{\text{train}}\) and \(D_{\text{train}}\) without extra manual supervision, on an \(A_{\text{test}}\) image captured at \(60^\circ\).
\section{Conclusion}
\label{sec:conclusion}
In this work, we presented practical workflows for training object detection models to detect and count German baked goods with limited supervision.
Although centered on a specific application, our work addresses the broader challenges of adapting computer vision models for industrial deployment: tasks are highly specialized, and annotated training data are scarce.
We discussed how bakeries can benefit from integrating computer vision models into their workflow, and demonstrated how to effectively tailor models to specific assortments of baked goods without relying on large, fully annotated datasets.
To train and evaluate our models, we compiled several datasets with varying levels of supervision. 
This included a fully annotated dataset split, used to train a strong baseline and serve as a benchmark for evaluation.
We combined the localization capabilities of open-vocabulary object detection models, OWLv2 and Grounding DINO, with image-level supervision to annotate a training set in a weakly supervised manner.
While this approach required limiting each image to a single class, it significantly reduced annotation costs.
Our results show that, while these open-vocabulary models, combined with minor post-processing, excel at localizing baked goods, they struggle to differentiate between them, likely due to our highly specialized domain.
Using weak supervision, we trained the one-stage object detection model YOLOv11, which achieved a mean average precision (mAP) of \(0.91\) on our benchmark, relying exclusively on image-level supervision.
We used Segment Anything 2 as a pseudo-label propagation model to annotate a training set of video frames, cutting annotation cost by over \(96\%\).
Fine-tuning with these pseudo-labeled data improved robustness to viewpoint variations and significantly increased performance on our more challenging video frame test set, boosting mAP by \(19.3\%\).
Importantly, these gains persisted when annotations were inferred automatically.
By combining weak and pseudo-supervised training, our trained models outperformed our baseline model trained in a fully supervised manner on our video frame test set, using only image-level supervision.
Our work exemplifies how to adapt strong pre-existing computer vision models, which lack domain-specific knowledge, for highly specialized tasks.
%
%
\printbibliography

\end{document}